\documentclass[journal]{IEEEtran}
\usepackage{graphicx}
\usepackage{subfigure}
\usepackage{float}
\usepackage{amsmath}
\usepackage{amssymb}
\usepackage{makecell}
\usepackage{caption}

\usepackage{graphicx}
\usepackage{tabu}
\usepackage{makecell}                 % 三线表-竖线
\usepackage{booktabs}                 % 三线表-短细横线
\usepackage{multirow}
\usepackage{pifont}
\usepackage{algorithm}
\usepackage{algorithmic}
\usepackage{amsfonts}
\usepackage{graphicx}
\usepackage{booktabs}
\usepackage{cleveref}
\usepackage{soul, color, xcolor}

\ifCLASSINFOpdf
\else
\fi
\begin{document}
	
	\title{Dynamic Degradation Decomposition Network for All-in-One Image Restoration}
	
	\author{Huiqiang Wang,
		Mingchen Song,
		and Guoqiang Zhong,~\IEEEmembership{Member,~IEEE}
		\thanks{
			This work was partially supported by the xxx. (Corresponding author: Guiqiang Zhong.)
			
			H. Wang, M. Song, and G. Zhong are with the College of Computer Science and Technology, Ocean University of China, Qingdao 266100, China (e-mail: wanghuiqiang0505@foxmail.com; mingchens0905@gmail.com; gqzhong@ouc.edu.cn). H. Wang and M. Song contributed equally to this work.
			
		}
	}
	%\thanks{Manuscript received Month Date, Year; revised Month Date, Year.}}


\markboth{Journal of \LaTeX\ Class Files, VOL. XXX, NO.XXX, MARCH 2025}
{Shell \MakeLowercase{et al.}: Huiqiang Wang}

\maketitle
\begin{abstract}
Currently, restoring clean images from a variety of degradation types using a single model is still a challenging task. Existing all-in-one image restoration approaches struggle with addressing complex and ambiguously defined degradation types. In this paper, we introduce a dynamic degradation decomposition network for all-in-one image restoration, named D$^3$Net. D$^3$Net achieves degradation-adaptive image restoration with guided prompt through cross-domain interaction and dynamic degradation decomposition. Concretely, in D$^3$Net, the proposed Cross-Domain Degradation Analyzer (CDDA) engages in deep interaction between frequency domain degradation characteristics and spatial domain image features to identify and model variations of different degradation types on the image manifold, generating degradation correction prompt and strategy prompt, which guide the following decomposition process. Furthermore, the prompt-based Dynamic Decomposition Mechanism (DDM) encourages the network to adaptively select restoration strategies utilizing the two-level prompt generated by CDDA. Thanks to the synergistic cooperation between CDDA and DDM, D$^3$Net achieves superior flexibility and scalability in handling unknown degradation, while effectively reducing unnecessary computational overhead. Extensive experiments on multiple image restoration tasks demonstrate that D$^3$Net outperforms the state-of-the-art approaches, especially improving PSNR by 5.47dB on SOTS-Outdoor and 3.30dB on GoPro, respectively.

\end{abstract}


%\ifCLASSOPTIONpeerreview
%\begin{center} \bfseries EDICS Category: 3-BBND \end{center}
%\fi
%\IEEEpeerreviewmaketitle

\section{Introduction}
\label{sec:intro}
\begin{figure}[tph!]  
\centering  
\includegraphics[width=0.5\textwidth]{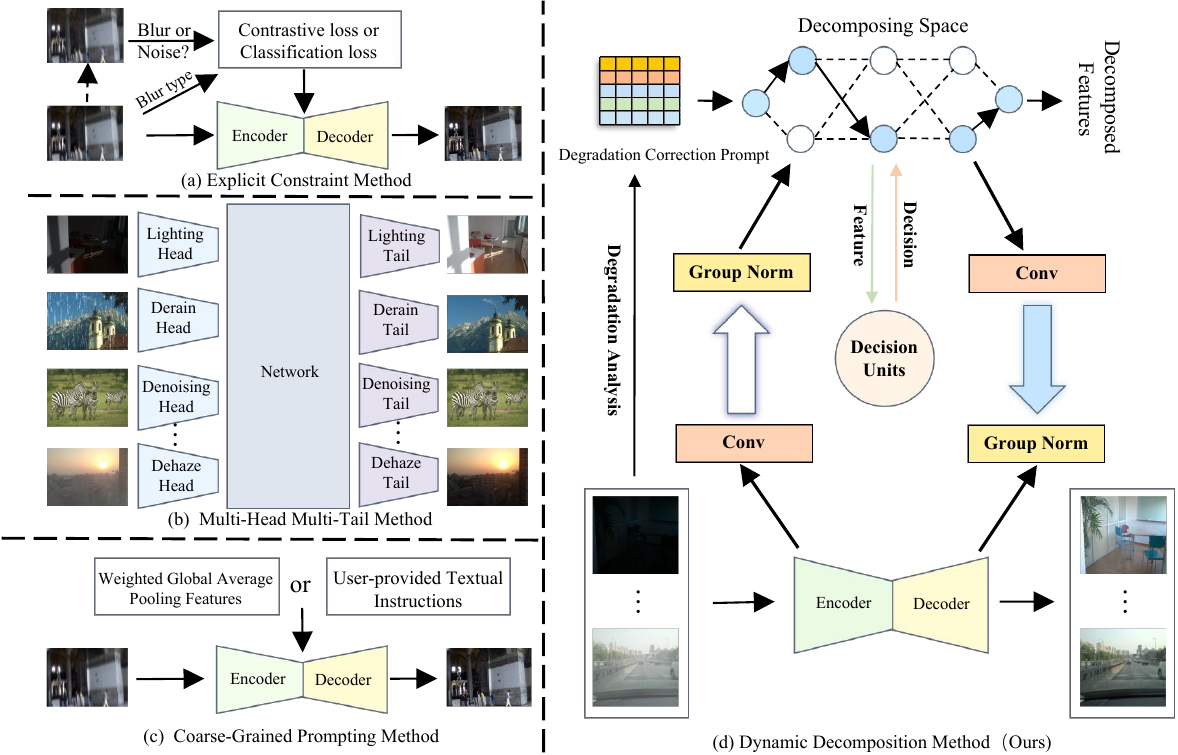}  
\caption{Conceptual differences between our work and other existing all-in-one methods (better viewed by zooming in).}  
\label{fig:1}  
\end{figure}  
\IEEEPARstart{I}{mage} restoration aims to remove degradations from corrupted images and generate high-quality ones. It is widely applied in various real-world scenarios, such as autonomous driving, digital photography, and security surveillance. Currently, most image restoration approaches focus on overcoming a single type of degradations, which may be denoising~\cite{Guo_Yan_Zhang_Zuo_Zhang_2019,Zhang_Zuo_Chen_Meng_Zhang_2017,Wang_Liu_Li_Han,yao2023towards,Zhang_Zhou_Jiang_Fu_2023}, deblurring~\cite{Asim_Shamshad_Ahmed_2018,Nah_Kim_Lee_2017,Pan_Ren_Hu_Yang_2018,Sree2023AdaptiveOD,Wu2023BroadSI}, deraining~\cite{Xiao2022ImageDT,Li2018RecurrentSC,Jiang2020MultiScalePF,Fu2016ClearingTS,Jiang2021MultiScaleHF}, dehazing~\cite{Engin2018CycleDehazeEC,Wu2021ContrastiveLF,Qu2019EnhancedPD,Zhao2021RefineDNetAW}, or low-light enhancement~\cite{Frants2023QCNNHSD,Ma2022TowardFF,Fu2023LearningAS,Xu2023LowLightIE,Wang2021LowLightIE,Xu2022SNRAwareLI}. Although these previous research has yielded promising results in a specific application, the challenge of addressing multiple degradation tasks in image restoration with a single model still persists. This significantly hampers the practical applicability of the existing image restoration techniques.

Recently, a series of all-in-one image restoration methods begin to attract widespread attention~\cite{Chen2020PreTrainedIP,Li2020AllIO,Li2022AllInOneIR,Zhang2023IngredientorientedML}. These studies aim to establish a unified model capable of addressing multiple degradation tasks. 
As shown in Fig.~\ref{fig:1}(a), the first category of these methods~\cite{Li2022AllInOneIR,Zhang2023IngredientorientedML} explores contrastive learning or explicit degradation classification constraints to establish a unified model for all-in-one image restoration.~Notably, during training, it is crucial to precisely differentiate degradation types in order to obtain positive and negative sample pairs.
This necessity may restrict its utility in practical scenarios where degradation types are often complex and not clearly defined.
Furthermore, as shown in Fig.~\ref{fig:1}(b), the second category of these methods seeks to build models using multiple independent heads and tails for all-in-one image restoration~\cite{Chen2020PreTrainedIP,Li2020AllIO}. However, this category of methods not only introduces an additional number of model parameters but also poses certain challenges in jointly optimizing multiple heads and tails, thereby reducing the scalability of the models.
Fig.~\ref{fig:1}(c) showcases the third category of methods, which guide all-in-one image restoration through the use of weighted global average pooling features or user-provided textual instructions~\cite{PromptIR_NIPS2024,conde2024instructir}.  While these approaches exhibit advantages by integrating semantic information, they are fundamentally limited by prompt ambiguity and coarse-grained guidance, directly restricting model performance in handling image restoration tasks with significant spatial variations.

To design a flexible and scalable all-in-one image restoration architecture, we propose two important research questions based on the limitations of the three aforementioned approaches:
i)~\textit{``How can we break through the limitations of traditional degradation feature extraction?"} Existing all-in-one image restoration methods primarily rely on spatial domain analysis, overlooking the rich degradation information inherent in frequency domain representations.  However, the frequency domain not only provides abundant signal features but also effectively reveals various types of degradation patterns. Therefore, developing a mechanism capable of generating and effectively utilizing frequency domain information becomes critical for enhancing the performance of all-in-one image restoration.
ii)~\textit{``How can we design a network architecture that dynamically adapts to complex degradation scenarios?"} Traditional static linear processing paradigms struggle to address the diverse range of degradation conditions. The key challenge lies in constructing a mechanism that enables the network to autonomously perceive degradation features, adjust computational pathways, as well as achieve efficient and smooth transitions between different degradation contexts.

As shown in Fig.~\ref{fig:1}(d), in this paper, we propose the Dynamic Degradation Decomposition Network (D$^3$Net) for all-in-one image restoration, which aims to address unknown multiple degradation situations. Unlike previous all-in-one image restoration approaches~\cite{Chen2020PreTrainedIP,Li2020AllIO,Li2022AllInOneIR,Zhang2023IngredientorientedML,PromptIR_NIPS2024,conde2024instructir}, our method departs from the perspective of \textit{frequency domain} and \textit{dynamic networks} to effectively handle unknown and complex degradations without relying on explicit degradation labels or predefined prompts.
In order to address the first question as above, we design a Cross-Domain Degradation Analyzer (CDDA) that generates degradation-related prompt to guide the degradation decomposition process. Specifically, it considers the complementary degradation characteristics between frequency and spatial domains, deeply interacting key degradation features in the frequency domain with spatial image features to identify and model the degradation types on the image manifold. Meanwhile, it generates degradation correction prompt and strategy prompt, thereby providing guidance for the following degradation decomposition process. Moreover, for the second question, we introduce a degradation-aware Dynamic Decomposition Mechanism (DDM). The design of DDM aims to progressively and dynamically decompose degradation features, fully leveraging the two-level prompt generated by CDDA to guide the image restoration process. Specifically, this mechanism distinctly enables the network to dynamically adapt the input from diverse decomposition blocks, tailoring its model to suit different degradation scenarios. 

In summary, our contributions are as follows:
\begin{itemize}
\item We propose a novel universal network to perform all-in-one image restoration tasks. The network achieves the restoration of degraded images by leveraging prompt information through adaptive cross-domain interaction and a dynamic decomposition mechanism.
\item We design a novel cross-domain degradation analyzer that generates degradation correction prompt and strategy prompt. It equivalently imposes effective two-level prompt constraints on the restored image manifold, thereby enhancing the accuracy and adaptability of the restoration process.

\item We introduce a novel dynamic decomposition mechanism. Owing to its adaptive prompt-guided mode, this mechanism can dynamically select an appropriate restoration strategy based on the degradation state of the images, thereby achieving considerable degradation decomposition effects while reducing computational overhead.
\item Extensive experiments on popular benchmark datasets demonstrate the effectiveness and scalability of D$^3$Net, which achieves the state-of-the-art performance on multiple image restoration tasks.
\end{itemize}

\section{Related Work}
In this section, we provide an overview of recent advancements relevant to the topics of this paper, including image restoration and dynamic networks. 
\subsection{Image Restoration}
\textbf{Image Restoration for Single Degradation.} Traditional image restoration techniques mainly target at a specific type of degradation, such as blur, noise, or haze. These approaches~\cite{R_Girish_Ambasamudram_2019,Xu_Zhang_Zhang_Feng_2017,Luo2015RemovingRF,Malhotra2016SingleIH,Dong2010ImageDA}, optimized for specific problems, can often achieve good results in the corresponding degradation scenario. For example, Mohan~et al.~\cite{R_Girish_Ambasamudram_2019} propose a model-based approach for addressing blur caused by camera shake. However, such approaches typically rely on specific prior degradation models, such as linear blur or motion blur. These model-based approaches may be severely limited when the actual degradation process does not match the assumed one. Recently, with the advancement of deep learning, numerous image restoration approaches based on deep learning have emerged  and performed excellently across various image restoration tasks, such as denoising~\cite{Guo_Yan_Zhang_Zuo_Zhang_2019,Zhang_Zuo_Chen_Meng_Zhang_2017,Wang_Liu_Li_Han,yao2023towards,Zhang_Zhou_Jiang_Fu_2023}, deblurring~\cite{Asim_Shamshad_Ahmed_2018,Nah_Kim_Lee_2017,Pan_Ren_Hu_Yang_2018,Sree2023AdaptiveOD,Wu2023BroadSI}, deraining~\cite{Xiao2022ImageDT,Li2018RecurrentSC,Jiang2020MultiScalePF,Fu2016ClearingTS,Jiang2021MultiScaleHF,10336721}, dehazing~\cite{Engin2018CycleDehazeEC,Wu2021ContrastiveLF,Qu2019EnhancedPD,Zhao2021RefineDNetAW}, and low-light enhancement~\cite{Frants2023QCNNHSD,Ma2022TowardFF,Ma2022TowardFF,Fu2023LearningAS,Xu2023LowLightIE,Wang2021LowLightIE,Xu2022SNRAwareLI}. However, constrained by model architectures designed for single tasks, they tend to perform exceptionally well in single degradation removal tasks but struggle to deal with other types of degradation.
\\
\textbf{Image Restoration for Multiple Degradations.} When considering multiple degradation factors, the problem becomes more complex as each type of degradation may require 
a special strategy. To address this challenge, researchers begin to explore the multi-task learning architecture that uses specific prior information~\cite{Chen2020PreTrainedIP,Li2020AllIO,Jose_Valanarasu_Yasarla_Patel_2022,Liu2022TAPETP}. For example, Chen~et al.~\cite{Chen2020PreTrainedIP} use a multi-head and multi-tail architecture to address multiple degradations. However, these approaches not only introduce additional model parameters but also increase the complexity of model training. Recently, several works have explored unified models for all-in-one image restoration tasks. Li~et al.~\cite{Li2022AllInOneIR} propose a method using contrastive learning to distinguish different types of degradation, requiring explicit degradation classification constraints. Zhang~et al.~\cite{Zhang2023IngredientorientedML} introduce a two-stage ingredient-oriented multi-degradation learning framework to reformulate degradation, but their approach necessitates the manual definition of degradation types and prior knowledge.

Prompt-based methods have also been introduced into all-in-one image restoration. For example, Potlapalli et al.~\cite{PromptIR_NIPS2024} propose a method that relies on global average pooling to generate prompt weights. The coarse-grained prompts generated by such global methods may struggle to capture local degradation features, making them insufficient to handle spatially varying degradations. Consequently, when dealing with complex degradation patterns that cannot be effectively summarized by global statistics, their restoration performance deteriorates.

Similarly, Conde et al.~\cite{conde2024instructir} propose a method utilizing predefined degradation texts to guide multiple degradation types. However, this approach does not deeply model the inherent degradation features within the images. By focusing solely on user-provided textual instructions without in-depth analysis of degradation patterns in the images, it may overlook critical details necessary for optimal restoration, especially when degradations are subtle or intertwined.

In contrast, our method leverages the proposed cross-domain degradation analyzer to facilitate deep interactions between frequency-domain degradation features and spatial-domain image features. This interaction enhances the model's capacity to effectively capture local and complex degradation patterns, thereby improving its ability to represent fine-grained spatial variations in degradation. 
\subsection{Dynamic Networks}
The research on dynamic networks has gradually garnered attention in the academic community, with the objective of reducing computational complexity while ensuring model performance. In recent years, researchers introduce various methodologies to achieve this goal. For example, ~\cite{Park2015BiglittleDN} introduces the groundbreaking BL-DNN approach, which concatenates two distinct deep CNNs with varying depths. When the softmax prediction accuracy of the first CNN surpasses a predefined threshold, the network terminates prematurely to reduce computational complexity. Subsequent works~\cite{Zhu_Han_Wu_Zhang_Nie_Lan_Wang_2021, Li_Wang_Wang_Liang_Li_Chang_2021, Wu_Nagarajan_Kumar_Rennie_Davis_Grauman_Feris_2018, Yu_Wang_Dong_Tang_Loy_2021} have also improved recognition tasks based on this concept.
Moreover, some researchers develop methods that dynamically adjust convolutional kernel weights~\cite{Yang2019CondConvCP,Chen2019DynamicCA,Harley2017SegmentationAwareCN,Su2019PixelAdaptiveCN} and kernel shapes~\cite{Dai2017DeformableCN,Zhu2018DeformableCV,Gao2019DeformableKA} during inference based on their inputs. However, due to their irregular memory access and computation patterns, these methods are limited in further efficiency improvement. Diverging from most existing dynamic networks, we design a degradation prompt-based dynamic decomposition mechanism that dynamically selects appropriate image restoration strategies.
\begin{figure*}[t] 
\centering 
\includegraphics[width=0.98\textwidth]{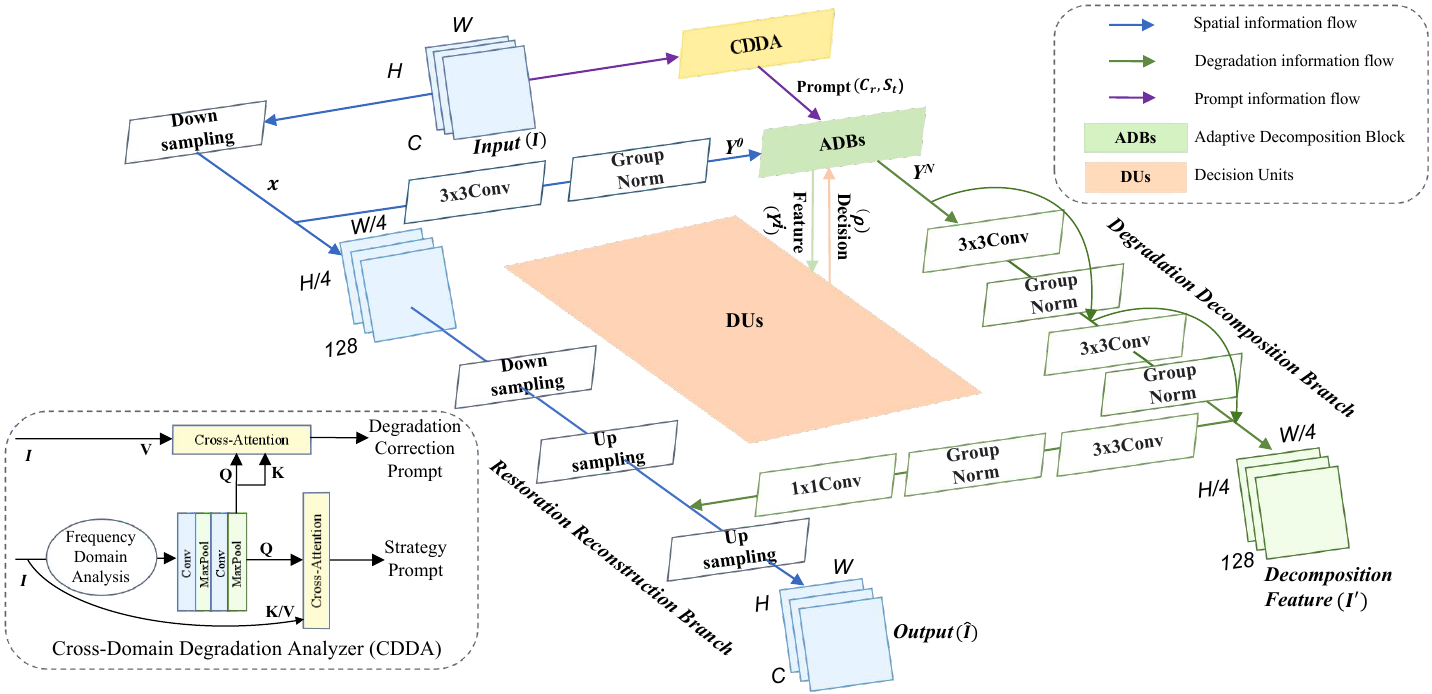} 
\caption{The architecture of D$^3$Net, which consists of a restoration reconstruction branch and a degradation decomposition branch.} 
\label{fig:2} 
\end{figure*}
\section{Method}
In this section, 
%we introduce the Dynamic Degradation Decomposition Network for Multi-Degradation Image Restoration, dubbed D3-Net. 
we outline the architecture of D$^3$Net in Section 3.1, while the proposed  cross-domain degradation analyzer and dynamic decomposition mechanism are presented in Sections 3.2 and 3.3, respectively.
\subsection{The Architecture of D$^3$Net}
In order to achieve prompt-guided adaptive image degradation restoration, we design a D$^3$Net as illustrated in Fig.~\ref{fig:2}. It primarily consists of a restoration reconstruction branch and a degradation decomposition branch. The restoration reconstruction branch employs a generic U-Net structure~\cite{Ronneberger_Fischer_Brox_2015} to reconstruct the input image. Building upon this foundational process, for a given degraded input image $I$, the shallow degradation features $x$ extracted by the first encoder of the restoration reconstruction branch are inputted into the degradation decomposition branch. That is, $x = f_{\text{encoder}}(I)$. Concurrently, $I$ is fed into the cross-domain degradation analyzer, which generates degradation correction prompt $C_t$ and strategy prompt $S_t$ to guide the degradation decomposition branch. 
Under the guidance of $C_t$ and $S_t$, it then iteratively decomposes the features through multiple stages. Specifically, at each stage $k$ ($k=1,2,...,N$), the branch selects degradation-adaptive strategies to decompose the current features $Y^{k-1}$ into refined degradation-decomposed $Y^k$ and an updated restoration strategy $\rho^k$, i.e., 
\[
Y^k, \rho^k = f_{\text{ddm}}(Y^{k-1}, C_t, S_t).
\]
After $N$ stages of decomposition, the final degradation-decomposed features $Y^N$ are passed through a feature refinement module to generate the decomposition-finalized features $I'$.
In D$^3$Net, we continuously feed $I'$ from the degradation decomposition branch back into the feature maps to enhance the restoration capability of the restoration reconstruction branch, i.e. $\hat{I} = f_{\text{rec}}(I, I')$. This feedback mechanism encourages both branches to collaborate in jointly managing image restoration and correction, thereby enhancing the overall quality of image restoration.

Additionally, D$^3$Net can be trained end-to-end without introducing additional degradation classification losses. It solely uses the $\mathcal{L}_1$ loss as the overall reconstruction loss, as shown in the following:
\begin{equation}
\mathcal{L}_1 = \|\hat{I} - I_{gt}\|_1,
\end{equation}
where $\hat{I}$ represents the reconstructed image and $I_{gt}$ represents the ground truth image.
\subsection{The Cross-Domain Degradation Analyzer}
CDDA aims to understand and alleviate the pervasive degradation problems that occur during the acquisition and transmission of images. This prevalent degradation manifests in various forms, including common issues such as noise, raindrops, haze, low-light conditions, motion, and focus blur, all contributing to pixel inconsistency and structural distortion in the spatial domain. 
For example, noise introduces random fluctuations in brightness, raindrops create localized bright spots and blurring, fog leads to reduced contrast and color shifts, low-light conditions affect brightness and visibility, and motion or focus blur results in unclear edges and details.
The complexity of these effects intertwines spatially and is content-dependent, posing a significant challenge for the separation and correction of image degradation. 

To address the interplay of these effects, CDDA employs Fourier transform analysis~\cite{harris1978use} to study the spectral characteristics of images. Specifically, the methodology proceeds as follows:
First, for each image $I$ in a batch, we convert its color space and normalize it to prepare for frequency domain analysis. The two-dimensional discrete Fourier transform of image $I$ is represented as:
\begin{equation}
F(u, v) = \sum_{x=0}^{m-1} \sum_{y=0}^{n-1} I(x, y) \cdot e^{-j2\pi\left(\frac{ux}{m} + \frac{vy}{n}\right)},
\end{equation}
where $m$ and $n$ are the number of pixels in the horizontal and vertical directions of the image, whilst $u$ and $v$ are the frequency domain coordinates. Additionally, to shift the low-frequency components to the center of the spectrum for analysis, we apply a frequency domain shift operation, obtaining:
\begin{equation}
%F_{\text{shift}}(u, v) = F(u, v) e^{j\pi(u+v)}.
F_{\text{shift}}(u, v) = F(u, v) \cdot e^{-j\pi(u+v)}.
\end{equation}
Subsequently, we compute the amplitude spectrum $M(u, v)$ to represent the energy distribution of the image:
\begin{equation}
M(u, v) = |F_{\text{shift}}(u, v)|.
\end{equation}
Here, the amplitude spectrum $M(u, v)$ provides us with a phase-independent view of the energy distribution, revealing the spectral distribution and areas of energy concentration. In this view, different degradation modes exhibit specific features: raindrops and haze tend to affect certain frequency ranges; raindrops cause horizontal stripes or diagonal stripes; haze usually results in a decrease in low-frequency components; noise typically manifests as high-frequency components in the frequency domain; blur is characterized by a significant weakening of high-frequency components; thus, this step helps us simplify the complexity of degradation in the spatial domain.

After extracting frequency domain features, we design a degradation correction prompt generation framework. Specifically, for the given frequency domain features $M(u,$ $v)$, we utilize a frequency domain analysis block to extract key information, where two $3 \times 3$ convolutional layers and two max-pooling layers are used to extract processed frequency domain degradation features $D_f$. Formally, this process can be represented as:
\begin{equation}
D_f = \text{Pool}(\text{ReLU}(\text{Pool}(\text{ReLU}(\text{Conv}({M}(u, v)))))).
\end{equation}
These processed frequency domain features $D_f$ reveal the key patterns of image degradation, providing a foundation for subsequent spatial domain feature interaction. Then,
we integrate frequency-domain degradation features $D_f$ with the original spatial domain image features $I$ to derive two distinct prompts: the degradation correction prompt $C_t$ and the strategy prompt $S_t$. The integration process is orchestrated via two specialized cross-attention mechanisms:
\begin{equation}
C_t = \text{CrossAttention}(Q_{I}, K_{I}, \phi(V_{D_f})),
\end{equation}
\begin{equation}
S_t = \text{CrossAttention}(\phi(Q_{I}), K_{D_f}, V_{D_f}).
\end{equation}
Here, $Q$ denotes the query, $K$ signifies the key, and $V$ represents the value, each originating from different feature sources, i.e., $D_f$ or $I$. In generating $C_t$, $I$ serves as both $Q$ and $K$, while $D_f$ acts as $V$, undergoing non-linear projection through $\phi(\cdot)$, which consists of three layers of $3 \times 3$ convolutions. This design ensures that $C_t$ is tightly aligned with spatial image features, emphasizing the fine-grained degradation details that need correction, while injecting frequency-based degradation prompt from $D_f$.
\begin{figure*}[t] 
\centering 
\includegraphics[width=0.95\textwidth]{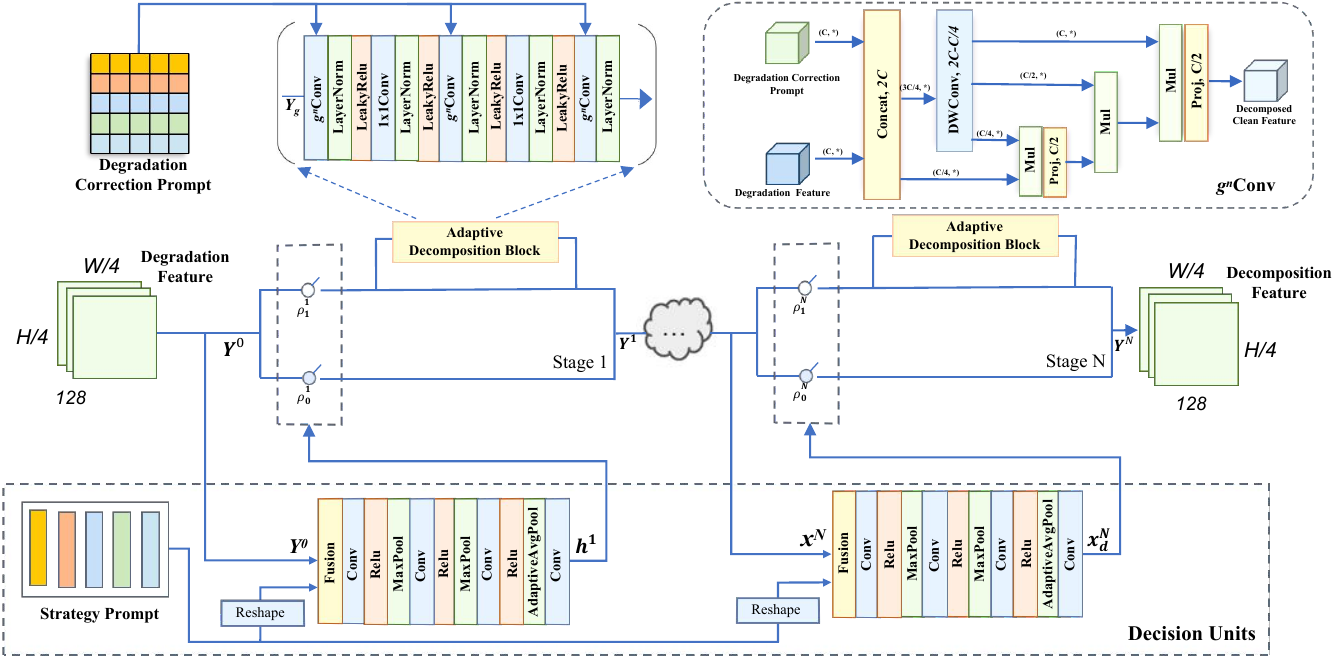} 
\caption{Dynamic decomposition mechanism: the upper part is adaptive decomposition blocks and the lower part is decision units.} 
\label{fig:ddm} 
\end{figure*}

Similarly, in generating $S_t$, $I$ is non-linearly projected into $Q$, and $D_f$ serves as both $K$ and $V$. This reflects that the global restoration strategy (captured by $S_t$) relies on frequency-domain insights to determine how to navigate multiple possible restoration paths, while still referencing the image content. This design integrates frequency domain features with spatial domain characteristics, effectively leveraging their complementarity in the image degradation analysis process. $C_t$ and $S_t$ form a two-level prompt that encapsulates fine-grained correction instructions and higher-level strategic guidance, respectively.

\subsection{Dynamic Decomposition Mechanism}
As shown in Fig.~\ref{fig:ddm}, the dynamic decomposition mechanism consists of two parts: Decision Units (DUs) and Adaptive Decomposition Blocks (ADBs).
Below, we introduce DUs and ADBs respectively.
\subsubsection{Decision Units} 
The goal of DUs is to enable the network to make decisions based on the input degradation features and the two-level prompt at the current processing stage. A DU consists of a multi-layer convolutional neural network structure, which can effectively compress the feature space through stacked convolutional layers and pooling layers, extracting independent hard decision vectors to ultimately determine whether to activate the corresponding ADB. Mathematically, the processing of a DU can be described as a function $\mathcal{F}_{DU}$, which maps input features to a decision space. For the input degraded feature map $Y^{i-1}$ at the $i$-th stage, with $Y^0$ being the initial degraded feature map, this mapping can be formalized as:
%For the input degraded feature map $Y^{i-1}$ at the $i$-th stage, this mapping can be formalized as:
\begin{equation}
	h^i = \mathcal{F}^{i}_{DU}(Y^{i-1}, S_t),
\end{equation}
where $h^i = [h^i_0, h^i_1] \in \mathbb{R}^2$ represents the output signal of the $i$-th DU, with $h^i_0$ and $h^i_1$ denoting the logits for suppressing or activating the corresponding ADB, respectively.
$S_t$ represents the strategy prompt.

To provide continuous gradients during training and achieve accurate discrete decisions during inference, we apply the Gumbel Softmax trick \cite{Jang_Gu_Poole_2016,Zhu_Han_Wu_Zhang_Nie_Lan_Wang_2021,Song_Chen_Zhang_2023}.
Its expression can be written as:
\begin{equation}
	\label{eq:gumbel_softmax}
	\rho^i_m = \frac{\exp((h^{i}_m + g^{i}_m)/\tau)}{\sum_{k=0}^{1} \exp((h^{i}_k + g^{i}_k)/\tau)}, \quad m = 0, 1,
\end{equation}
where \( \rho^i_m \) denotes the probability of selecting option \( m \) for the \( i \)-th decision unit, with \( m = 0 \) for suppression and \( m = 1 \) for activation. Here, $g^{i}_m$ is independent and identically distributed (i.i.d.) noise sampled from the Gumbel distribution for the $m$-th option, and $\tau$ is a temperature parameter controlling the smoothness of the approximation.
This function approximates the sampling process of a discrete distribution while maintaining differentiability, facilitating end-to-end training.

Importantly, the DUs and $S_t$ work together: while $S_t$ encodes a global restoration strategy derived from cross-domain interactions, the DUs use this strategic knowledge to adaptively choose which decomposition stages to apply, ensuring efficient and targeted feature refinement.

After a DU outputs a decision, each ADB dynamically adjusts its operations according to the decision of the DU. Specifically, each ADB is activated or suppressed by an independent control signal. 
This interaction between each ADB and its corresponding DU forms a distinct processing stage in the overall system. The operation of each stage is governed by the decision dynamics between the DU and the ADB, ensuring that only relevant feature transformations are applied using a differentiable gating mechanism. This can be expressed as the following formula:
\begin{equation}
Y^{i} = Y^{i-1} + \mathcal{F}_{ADB}^{i}(Y^{i-1}) \odot \rho^i_1, \quad i = 1, \ldots, N,
\end{equation}
where $Y^{i}$ is the output after the $i$-th stage, $Y^{i-1}$ is the input feature from the previous stage, $\mathcal{F}_{ADB}^{i}(\cdot)$ denotes the $i$-th ADB’s operations, and $\rho^i_1$ is the activation probability obtained from the Gumbel-Softmax corresponding to $h^i_1$, serving as a differentiable gating mechanism that controls the contribution of the ADB's output.
We anneal the temperature $\tau$ from 1.0 to 0.1 over the training process, encouraging $\rho^i_1$ to approach a near-discrete distribution (close to 0 or 1) while remaining differentiable.
At inference time, we compute $\rho^i_1$ using the deterministic Softmax formulation in Eq.~\ref{eq:gumbel_softmax} with $g^i_m = 0$, ensuring stable decision-making without gradient computation.
$N$ is the number of DUs, which is set to 12 in our work.
In this way, our DDM ensures stable gradient transmission during end-to-end training while achieving highly flexible and dynamic adjustments.

% TABLE
\subsubsection{Adaptive Decomposition Blocks}
Having briefly mentioned ADBs earlier, this section delves deeper into their operational mechanics, especially the adaptive decomposition of pixel-level degradation features through the utilization of the degradation correction prompt. To achieve this, we design a module that combines the input degradation feature map with the degradation correction prompt, which constitutes the basic component of the dynamic decomposition mechanism proposed in this paper. Specifically, each ADB consists of five sub-blocks: two convolutional blocks containing \( g^n \)Conv~\cite{Rao_Zhao_Tang_Zhou_Lim_Lu_2022}, interleaved with two \( 1 \times 1 \) convolution blocks, and an output layer. Notably, we have improved \( g^n \)Conv. Specifically, $C_t$ is integrated as a direct input to each $g^n$Conv layer alongside the degradation feature $Y_{g}$, ensuring that the decomposition operation is informed by explicit correction instructions at each pixel:
\begin{equation}
\begin{split}
	[p_0^{HW \times C}, q_0^{HW \times C}] = \phi_{\text{concat}}([{Y_{g}}, C_t]) &\in \mathbb{R}^{HW \times 2C},\\
\end{split}
\end{equation}
where $Y_{g}\in \mathbb{R}^{HW \times C} $  is the degraded features to be decomposed, and $\phi_{\text{concat}}$ concatenates these features and the degradation correction prompt $ C_t \in \mathbb{R}^{HW \times C} $ along the channel dimension. This change not only enhances the information flow between channels but also allows the degradation correction prompt to directly participate in the adjustment of each feature. Next, a depthwise separable convolution $f$ is employed to model local spatial relationships and perform degradation corrections:
\begin{equation}
{p}_1 = f({q}_0) \odot {p}_0 \in \mathbb{R}^{HW \times C},
\end{equation}
Although $f$ remains a depthwise separable convolution, the change in input enables $f$ to capture richer contextual information.
Through this method, we realize the first-order interaction between degradation features and dynamic correction prompt, which can be further expressed by the following formula:
\begin{equation}
p_1^{(i, c)} = \sum_{j \in \Omega_i} w_{i \rightarrow j}^c q_0^{(j, c)} p_0^{(i, c)},
\end{equation}
where \( \Omega_i \) is a local window centered on pixel \( i \), and \( w \) represents the weight of the depthwise separable convolution \( f \). This interaction not only focuses on the features of each pixel itself but also includes the features of neighboring pixels. In other words, this design allows the degradation correction prompt and degradation features to interact at each pixel level, thereby achieving adaptive degradation decomposition. 
The final output \( {Y_g^j} \) is generated by another linear layer \( \phi_{\text{out}} \):
\begin{equation}
{Y_g^j} = \phi_{\text{out}}({p}_1) \in \mathbb{R}^{HW \times C}, \quad j = 1, 2, 3.
\end{equation}
Here, \(j\) denotes the order of the processing using the \(g^n\)Conv module at the current stage. 

In summary, unlike traditional image restoration methods that struggle to handle multiple degradation types, D$^3$Net separates the restoration process into degradation correction and restoration strategy guidance, using the proposed CDDA and DDM respectively. This synergy enables D$^3$Net to efficiently adapt to the nuances of degradation, improving the overall restoration quality without incurring the computational burden. The integration of frequency-domain insights into the restoration process provides a deeper understanding of degradation patterns, allowing for accurate and targeted restoration. Moreover, D$^3$Net's unique approach to dynamic feature refinement, driven by guided prompts, ensures that restoration strategies are dynamically tailored to specific degradation characteristics, offering a high level of precision and effectiveness across a variety of image restoration tasks.

\section{Experiments}
In this section, we first introduce the experimental settings, followed by the qualitative and quantitative comparison results with the state-of-the-art (SOTA) approaches. Lastly, we report the results of ablation studies.
\begin{table*}[t]
\renewcommand{\arraystretch}{1.2}
\setlength{\tabcolsep}{4pt}
\centering
\caption{Quantitative results on five challenging image restoration datasets. The best results are highlighted in \textbf{bold}.}
%    \resizebox{\linewidth}{!}{
	\scalebox{1}{
		\begin{tabular*}{\linewidth}{c|cc|cc|cc|cc|cc|cc|c|c}
			\Xhline{1pt}\rule{0pt}{10pt}
			
			\multirow{2}{*}{\textbf{Method} }
			& \multicolumn{2}{c|}{Rain100L~\cite{Yang_Tan_Feng_Liu_Guo_Yan_2017}} & \multicolumn{2}{c|}{SOTS~\cite{Li_Ren_Fu_Tao_Feng_Zeng_Wang_2019} }  &\multicolumn{2}{c|}{BSD68~\cite{Martin_Fowlkes_Tal_Malik_2002} } & \multicolumn{2}{c|}{GoPro~\cite{Nah_Kim_Lee_2017}}& \multicolumn{2}{c|}{LOL~\cite{Wei_Wang_Yang_Liu_2018}}& \multicolumn{2}{c|}{Average}
			%				&\multirow{2}{*}{Params/FLOPs}
			&\multirow{2}{*}{Params}
			&\multirow{2}{*}{FLOPs}
			\\
			
			%\cmidrule(lr){2-3}\cmidrule(lr){4-5}\cmidrule(lr){8-10}\cmidrule(lr){11-13}
			& PSNR↑  & SSIM↑ & PSNR↑  & SSIM↑ & PSNR↑  & SSIM↑  & PSNR↑  & SSIM↑& PSNR↑  & SSIM↑ & PSNR↑  & SSIM↑ & & \\
			
			\hline \rule{0pt}{10pt}
			NAFNet~\cite{Chen_Chu_Zhang_Sun_2022}     & 35.56 & 0.967 & 25.23 & 0.939 & 31.02 & 0.883 & 26.53 & 0.808 & 20.49 & 0.809 & 27.76 & 0.881 & 17.11M&63.28G\\
			HINet~\cite{Chen_Lu_Zhang_Chu_Chen_2021}      & 35.67 & 0.969 & 24.74 & 0.937 & 31.00 & 0.881 & 26.12 & 0.788 & 19.47 & 0.800 & 27.40 & 0.875 & 88.67M&170.52G\\
			MPRNet~\cite{Zamir_Arora_Khan_Hayat_Khan_Yang_Shao_2021}    & 38.16 & \textbf{0.981} & 24.27 & 0.937 & 31.35 & \textbf{0.889} & 26.87 & 0.823 & 20.84 & 0.824 & 28.27 & 0.890 & 15.74M&426.84G\\
			DGUNet~\cite{Mou_Wang_Zhang_2022}    & 36.62 & 0.971 & 24.78 & 0.940 & 31.10 & 0.883 & 27.25 & 0.837 & 21.87 & 0.823 & 28.32 & 0.891 & 17.33M&216.54G\\
			MIRNetV2~\cite{Zamir2022LearningEF}  & 33.89 & 0.954 & 24.03 & 0.927 & 30.97 & 0.881 & 26.30 & 0.799 & 21.52 & 0.815 & 27.34 & 0.875 & 5.86M&35.23G\\
			SwinIR~\cite{Liang2021SwinIRIR}    & 30.78 &0.923 & 21.50 & 0.891 & 30.59 & 0.868 & 24.52 & 0.773 & 17.81 & 0.723 & 25.04 & 0.835 & 0.91M&43.76G\\
			Restormer~\cite{Zamir_Arora_Khan_Hayat_Khan_Yang_2022} & 34.81 & 0.962 & 24.09 & 0.927 & 31.49 & 0.884 & 27.22 & 0.829 & 20.41 & 0.806 & 27.60 & 0.881 & 26.13M&35.25G\\ \hline \rule{0pt}{10pt}
			DL~\cite{Fan_Chen_Yuan_Hua_Yu_Chen_2021}        & 21.96 & 0.762 & 20.54 & 0.826 & 23.09 & 0.745 & 19.86 & 0.672 & 19.83 & 0.712 & 21.05 & 0.743 & 2.09M&22.85G\\
			Transweather~\cite{Jose_Valanarasu_Yasarla_Patel_2022} & 29.43 & 0.905 & 21.32 & 0.885 & 29.00 & 0.841 & 25.12 & 0.757 & 21.21 & 0.792 & 25.22 & 0.836 & 37.93M&6.13G\\
			TAPE~\cite{Liu2022TAPETP}      & 29.67 & 0.904 & 22.16 & 0.861 & 30.18 & 0.855 & 24.47 & 0.763 & 18.97 & 0.621 & 25.09 & 0.801 & 1.07M&14.10G\\
			AirNet~\cite{Li2022AllInOneIR}    & 32.98 & 0.951 & 21.04 & 0.884 & 30.91 & 0.882 & 24.35 & 0.781 & 18.18 & 0.735 & 25.49 & 0.846 & 8.93M&75.32G\\
			PromptIR~\cite{PromptIR_NIPS2024}  & 35.14 & 0.959 & 24.16 & 0.913 & 31.27 & 0.885 & 26.57 & 0.825 & 21.26 & 0.829 & 27.57 & 0.881 &32.97M &370.64G\\ 
			IDR~\cite{Zhang2023IngredientorientedML}  & 35.63 & 0.965 & 25.24 & 0.943 & 31.60 & 0.887 & 27.87 & 0.846 & 21.34 & 0.826 & 28.34 & 0.893 & 15.34M&---\\ 
			
			InstructIR~\cite{conde2024instructir}  & 36.84 & 0.973 & 27.10 & 0.956 & 31.40 & 0.887 & 29.40 & 0.886 & 23.00 & 0.836 & 29.55 & 0.907 & 15.84M & 37.95G\\ 
			D$^3$Net (Ours)  & \textbf{38.35} & 0.973 & \textbf{32.57} & \textbf{0.965} & \textbf{31.73} & 0.860 & \textbf{32.70} & \textbf{0.851} & \textbf{26.49} & \textbf{0.857} & \textbf{32.36} & \textbf{0.901} & 
			37.80M&33.67G\\ 
			\Xhline{1pt}
		\end{tabular*}
		
	}
	
	\label{tab:allinone}
\end{table*}
\begin{figure*}[t] 
	\centering 
	\includegraphics[width=1\textwidth]{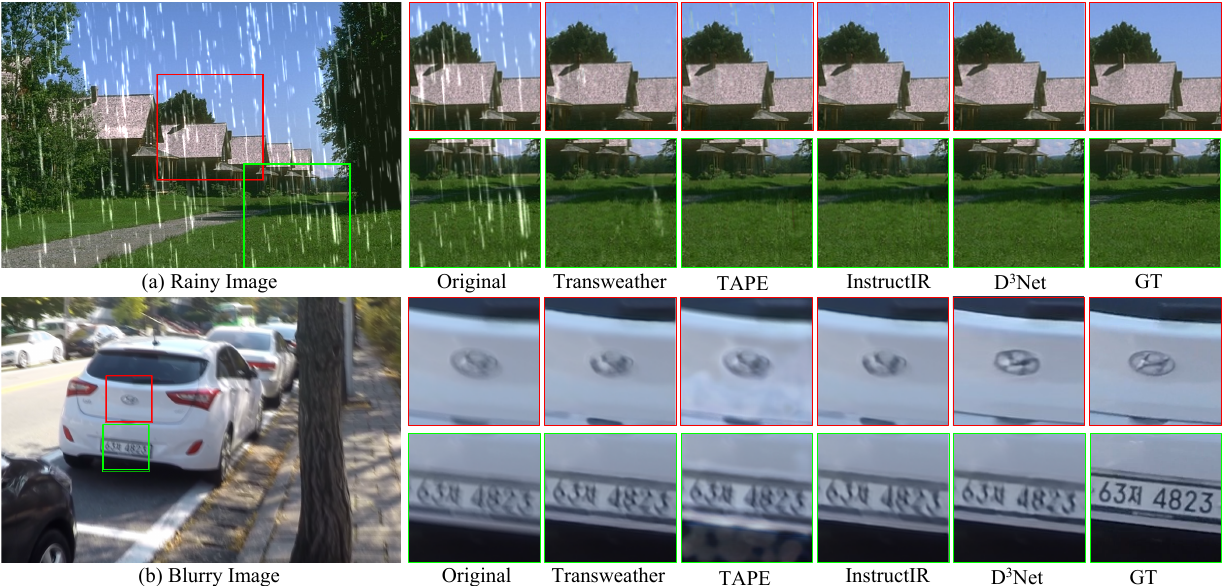} 
	\caption{Qualitative comparison with the SOTA methods on the Rain100L, and GoPro datasets (better viewed by zooming in).} 
	\label{fig:derainAndBlur} 
\end{figure*}
\begin{figure*}[t] 
	\centering 
	\includegraphics[width=0.99\textwidth]{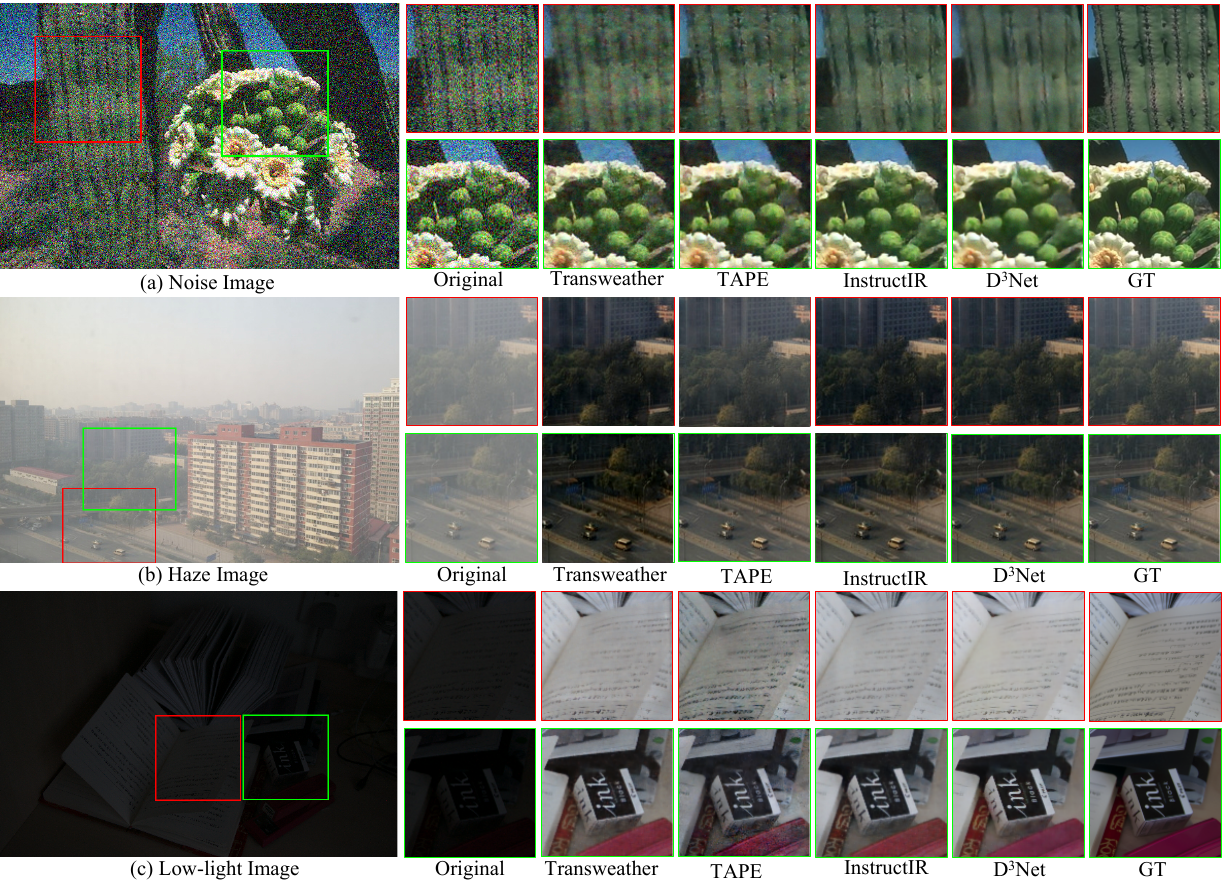} 
	\caption{Qualitative comparison with the SOTA methods on the BSD68, SOTS, and LOL datasets (better viewed by zooming in).} 
	\label{fig:threeTasks} 
\end{figure*}
\subsection{Experimental Settings}
\textbf{Datasets.} 
%This study aims to address the challenge of image restoration for multiple types of degradation. 
For a fair assessment of our proposed architecture under various degradation conditions, we construct a comprehensive dataset following~\cite{Zhang2023IngredientorientedML}, which integrates several well-recognized standard datasets, covering five important tasks in the field of image restoration. Specifically, the training set for D$^3$Net and its comparative methods includes: Rain200L~\cite{Yang_Tan_Feng_Liu_Guo_Yan_2017} for deraining, RESIDE~\cite{Li_Ren_Fu_Tao_Feng_Zeng_Wang_2019} for dehazing, BSD400~\cite{Martin_Fowlkes_Tal_Malik_2002} and WED~\cite{Ma_Duanmu_Wu_Wang_Yong_Li_Zhang_2017} for denoising, GoPro~\cite{Nah_Kim_Lee_2017} for deblurring, and LOL~\cite{Wei_Wang_Yang_Liu_2018} for low-light enhancement. 
%	Noteworthy, for the task of all-in-one image restoration in our study, these datasets need to be used simultaneously during the training process. 
For the task of all-in-one image restoration in our study, these datasets need to be used simultaneously during the training process.
\\
\textbf{Experimental Details.} We implement our model using the PyTorch framework and train it on four NVIDIA Geforce RTX 4090 GPUs. The model employs the Adam optimizer with exponential decay rates of $\beta_1 = 0.9$ and $\beta_2 = 0.999$, trained for 1200 epochs. The initial learning rate is set to $1e^{-4}$ and reduces to $1e^{-6}$ through a cosine annealing strategy. During training, $128\times128$ patches are randomly cropped from the original images as network inputs.
\\
\textbf{Performance Evaluation.} For performance evaluation, we employ Peak Signal-to-Noise Ratio (PSNR)~\cite{Huynh-Thu_Ghanbari_2008} and Structural Similarity Index (SSIM)~\cite{Wang_Bovik_Sheikh_Simoncelli_2004} as the primary metrics.
The selection of the test set follows the same diversity principle as the training set, using corresponding parts of the datasets including Rain100L~\cite{Yang_Tan_Feng_Liu_Guo_Yan_2017}, SOTS-Outdoor~\cite{Li_Ren_Fu_Tao_Feng_Zeng_Wang_2019}, BSD68~\cite{Martin_Fowlkes_Tal_Malik_2002}, GoPro~\cite{Nah_Kim_Lee_2017}, and LOL~\cite{Wei_Wang_Yang_Liu_2018}.
% Higher value of these metrics indicates better performance of the methods.

\begin{table}[t]
	\renewcommand{\arraystretch}{1.0}
	\setlength{\tabcolsep}{3pt}
	\footnotesize
	\centering
	\caption{Quantitative results of image denoising on the BSD68, Urban100, and Kodak24 datasets in terms of PSNR↑. The best results are highlighted in~\textbf{bold}.}       
	\scalebox{0.89}{
		\begin{tabular}{c|ccc|ccc|ccc}
			\Xhline{1pt}\rule{0pt}{10pt}
			
			\multirow{2}{*}{\textbf{Method} }
			& \multicolumn{3}{c|}{BSD68~\cite{Martin_Fowlkes_Tal_Malik_2002} } & \multicolumn{3}{c|}{Urban100~\cite{Huang_Singh_Ahuja_2015} }  &\multicolumn{3}{c}{Kodak24~\cite{Kodak2024} }
			\\
			
			%\cmidrule(lr){2-3}\cmidrule(lr){4-5}\cmidrule(lr){8-10}\cmidrule(lr){11-13}
			& $\sigma$=15  & $\sigma$=25 & $\sigma$=50  & $\sigma$=15  & $\sigma$=25 & $\sigma$=50& $\sigma$=15  & $\sigma$=25 & $\sigma$=50   \\
			
			\hline \rule{0pt}{10pt}
			%ProtoNet+FWT~\cite{Tseng_Lee_Huang_Yang_2020}(RN10)&-&23.77 & 26.87&-&38.87&43.78&-&67.34&75.74&-&72.72&85.82 \\
			NAFNet~\cite{Chen_Chu_Zhang_Sun_2022}     & 33.67  &31.02  &27.73 & 33.14 & 30.64  &27.20  &34.27  &31.80  &28.62\\
			HINet~\cite{Chen_Lu_Zhang_Chu_Chen_2021}  &33.72 &31.00 &27.63 &33.49 &30.94 &27.32 &34.38 &31.84&28.52 \\
			MPRNet~\cite{Zamir_Arora_Khan_Hayat_Khan_Yang_Shao_2021}&34.01& 31.35& 28.08 &34.13 &31.75 &28.41 &34.77 &32.31 &29.11\\
			DGUNet~\cite{Mou_Wang_Zhang_2022}&33.85 &31.10 &27.92& 33.67 &31.27& 27.94& 34.56& 32.10& 28.91\\
			MIRNetV2~\cite{Zamir2022LearningEF}&33.66& 30.97& 27.66& 33.30& 30.75& 27.22& 34.29& 31.81& 28.55\\
			SwinIR~\cite{Liang2021SwinIRIR}&33.31 &30.59& 27.13& 32.79& 30.18& 26.52& 33.89& 31.32& 27.93\\
			Restormer~\cite{Zamir_Arora_Khan_Hayat_Khan_Yang_2022}&34.03& 31.49& 28.11& 33.72 &31.26& 28.03& 34.78& 32.37& 29.08\\ 
			\hline \rule{0pt}{10pt}
			DL~\cite{Fan_Chen_Yuan_Hua_Yu_Chen_2021} &23.16  &23.09  &22.09  &21.10  &21.28 & 20.42 & 22.63  &22.66 & 21.95\\
			Transweather~\cite{Jose_Valanarasu_Yasarla_Patel_2022}&31.16 &29.00 &26.08 &29.64 &27.97 &26.08 &31.67 &29.64 &26.74 \\
			TAPE~\cite{Liu2022TAPETP}   & 32.86& 30.18& 26.63 &32.19& 29.65& 25.87& 33.24 &30.70 &27.19 \\
			AirNet~\cite{Li2022AllInOneIR} &33.49& 30.91& 27.66& 33.16 &30.83 &27.45& 34.14 &31.74 &28.59\\
			PromptIR~\cite{PromptIR_NIPS2024}&33.87 &31.27 &28.03 &33.52& 31.14 &27.91 &34.65 &32.13 &28.85\\ 
			IDR~\cite{Zhang2023IngredientorientedML}&\textbf{34.11} &31.60 &28.14 &33.82& 31.29 &28.07 &34.78 &32.42 &29.13\\ 
			InstructIR~\cite{conde2024instructir}&34.00 &31.40 &28.15 &33.77& 31.40 &28.13 &34.70 &32.26 &29.16\\ 
			D$^3$Net (Ours) &33.30 &\textbf{31.73} &\textbf{29.05} &\textbf{34.65} &\textbf{32.06} &\textbf{29.19} &\textbf{35.07} &\textbf{32.89} &\textbf{30.38} \\ 
			\Xhline{1pt}
		\end{tabular}
		
	}
	\label{tab:materialDenoise}
\end{table}
\begin{table}[t]
	\renewcommand{\arraystretch}{1.0}
	\setlength{\tabcolsep}{2pt}
	\small
	\centering
	
	\caption{Quantitative results on ~the TOLED and POLED datasets. The best results are highlighted in \textbf{bold}.}       
	%		\resizebox{\linewidth}{!}{
		\scalebox{0.83}{
			\begin{tabular}{c|ccc|ccc}
				\Xhline{1pt}\rule{0pt}{10pt}
				
				\multirow{2}{*}{\textbf{Method} }
				& \multicolumn{3}{c|}{TOLED~\cite{Zhou_Ren_Emerton_Lim_Large_2021}} & \multicolumn{3}{c}{POLED~\cite{Zhou_Ren_Emerton_Lim_Large_2021} }
				\\
				
				%\cmidrule(lr){2-3}\cmidrule(lr){4-5}\cmidrule(lr){8-10}\cmidrule(lr){11-13}
				& PSNR↑  & SSIM↑ &LPIPS↓ & PSNR↑  & SSIM↑ &LPIPS↓  \\
				
				\hline \rule{0pt}{10pt}
				%ProtoNet+FWT~\cite{Tseng_Lee_Huang_Yang_2020}(RN10)&-&23.77 & 26.87&-&38.87&43.78&-&67.34&75.74&-&72.72&85.82 \\
				NAFNet~\cite{Chen_Chu_Zhang_Sun_2022} & 26.89 &0.774 & 0.346 & 10.83 & 0.416 & 0.794 \\
				HINet~\cite{Chen_Lu_Zhang_Chu_Chen_2021} & 13.84 & 0.559 & 0.448 & 11.52 & 0.436 & 0.831 \\
				MPRNet~\cite{Zamir_Arora_Khan_Hayat_Khan_Yang_Shao_2021} & 24.69 & 0.707 & 0.347 & 8.34 & 0.365 & 0.798 \\
				DGUNet~\cite{Mou_Wang_Zhang_2022} & 19.67 & 0.627 & 0.384 & 8.88 & 0.391 & 0.810 \\
				MIRNetV2~\cite{Zamir2022LearningEF} & 21.86 & 0.620 & 0.408 & 10.27 & 0.425 & 0.722 \\
				SwinIR~\cite{Liang2021SwinIRIR} & 17.72 & 0.661 & 0.419 & 6.89 & 0.301 & 0.852 \\
				Restormer~\cite{Zamir_Arora_Khan_Hayat_Khan_Yang_2022} & 20.98 & 0.632 & 0.360 & 9.04 & 0.399 & 0.742 \\
				\hline \rule{0pt}{10pt}
				DL~\cite{Fan_Chen_Yuan_Hua_Yu_Chen_2021} & 21.23 & 0.656 & 0.434 & 13.92 & 0.449 & 0.756 \\
				Transweather~\cite{Jose_Valanarasu_Yasarla_Patel_2022} & 25.02 & 0.718 & 0.356 & 10.46 & 0.422 & 0.760 \\
				TAPE~\cite{Liu2022TAPETP}     &17.61 & 0.583 & 0.520 & 7.90 & 0.219 & 0.799\\
				AirNet~\cite{Li2022AllInOneIR}  & 14.58 & 0.609  &0.445 & 7.53 & 0.350 & 0.820 \\
				PromptIR~\cite{PromptIR_NIPS2024} & 26.53 &0.751 & 0.356 & 14.37  &0.461  &0.749 \\ 
				IDR~\cite{Zhang2023IngredientorientedML} & 27.91  &0.795 & 0.312 & 16.71  &0.497  &0.716 \\ 
				D$^3$Net (Ours)  &\textbf{29.05}    &\textbf{0.801}   &\textbf{0.223}   &\textbf{16.87}   &\textbf{0.503} &\textbf{0.639}\\ 
				\Xhline{1pt}
			\end{tabular}
			
		}
		
		\label{tab:materialUdc}
	\end{table}
	\subsection{Comparison with the SOTA approaches}
	We conduct a comprehensive evaluation by applying D$^3$Net to five complex image restoration scenarios, including deraining, dehazing, denoising, deblurring, and low-light image enhancement. 
	%In this study, the performance of D3-Net is qualitatively and quantitatively compared with seven general image restoration approaches and four all-in-one approaches.
	%\subsubsection{Quantitative Analysis}
	\\
	\textbf{Quantitative Analysis.} 
	We present comparative experiments between D$^3$Net and various general image restoration methods, such as NAFNet~\cite{Chen_Chu_Zhang_Sun_2022}, HINet~\cite{Chen_Lu_Zhang_Chu_Chen_2021}, MPRNet~\cite{Zamir_Arora_Khan_Hayat_Khan_Yang_Shao_2021}, DGUNet~\cite{Mou_Wang_Zhang_2022}, MIRNetV2~\cite{Zamir2022LearningEF}, SwinIR~\cite{Liang2021SwinIRIR}, and Restormer~\cite{Zamir_Arora_Khan_Hayat_Khan_Yang_2022}, as well as specialized all-in-one fashion methods like DL~\cite{Fan_Chen_Yuan_Hua_Yu_Chen_2021}, Transweather~\cite{Jose_Valanarasu_Yasarla_Patel_2022}, TAPE~\cite{Liu2022TAPETP}, AirNet~\cite{Li2022AllInOneIR}, PromptIR~\cite{PromptIR_NIPS2024}, IDR~\cite{Zhang2023IngredientorientedML}, and InstructIR~\cite{conde2024instructir}, across five image degradation tasks. As shown in Table~\ref{tab:allinone}, our D$^3$Net achieves the best performance in quantitative assessments across multiple datasets, specifically being equivalent to or surpassing the SOTA methods in terms of PSNR and SSIM metrics on datasets such as Rain100L, SOTS, BSD68, GoPro, and LOL. Particularly on the SOTS and LOL datasets, D$^3$Net achieves an improvement in PSNR by 5.47dB and 3.30dB, respectively, apparently outperforming other SOTA methods.

	However, it is observed that on the BSD68 and Rain100L datasets, the SSIM scores of D$^3$Net do not entirely surpass all other methods. This might be due to SSIM inclination towards evaluating structural similarity, whereas in some complex degradation scenarios, the restoration of overall structure may not be as critical as detail restoration. Moreover, considering the significant performance improvements achieved by D$^3$Net across multiple tasks, its number of parameters and FLOPs are reasonable.
	
	In addition, as shown in Table~\ref{tab:materialDenoise}, D$^3$Net exhibits performance that surpasses the SOTA methods on the BSD68~\cite{Martin_Fowlkes_Tal_Malik_2002}, Urban100~\cite{Huang_Singh_Ahuja_2015}, and Kodak24~\cite{Kodak2024} datasets under various noise levels (\(\sigma = 15, 25, 50\)). Particularly at a noise level of \(\sigma=50\), the performance improvement of D\(^3\)Net becomes notably significant, demonstrating its enhanced robustness in handling high-noise images. 
	
	In order to assess the model's capability in handling unknown and complex degradation, as illustrated in Table~\ref{tab:materialUdc}, we evaluate the restoration effects of our method compared to the SOTA methods in under-display camera (UDC) image restoration scenarios without any fine-tuning. The UDC problem holds significant research value in complex degradation analysis. Specifically, images captured under the UDC system exhibit complex degradation due to the point spread function and lower transmittance~\cite{Zhou_Ren_Emerton_Lim_Large_2021}. Therefore, evaluating such challenging and complex tasks allows us to better understand the generalization capabilities of the compared methods. Especially, our D$^3$Net shows superior performance on all evaluation metrics compared to the SOTA methods.
	%Furthermore, the parameter count and FLOPs suggest that D$^3$Net achieves great performance improvements in comparison to its resource investment.
	\\
	\textbf{Qualitative Analysis.} We conduct a series of qualitative experiments against the SOTA approaches to demonstrate the effectiveness of our proposed method. As shown in Figs.~\ref{fig:derainAndBlur} and~\ref{fig:threeTasks}, compared to other methods, D$^3$Net achieves stable improvements in visual perceptual quality on multiple degradation tasks.
	\subsection{Ablation Experiments}
	In this section, we conduct a series of ablation experiments on our proposed cross-domain interaction methods, dynamic decomposition mechanism, and decision units.
	\subsubsection{Ablation Study of D$^3$Net Components}
	We conduct a series of ablation experiments, as shown in Table~\ref{tab:ablation}. We sequentially remove Fourier Transform Analysis (FA), Cross-Attention Interaction (CA), Degradation Correction Prompt (DCP), Strategy Prompt (SP), and Decision Units (DUs), assessing the influence of each component on the performance of D$^3$Net. We report the average PSNR and SSIM values of different methods on five image degradation tasks. These results lead us to four key observations:
	\begin{table}[t]
		\renewcommand{\arraystretch}{1}
		\setlength{\tabcolsep}{7.3pt}
		\small
		\centering
		\caption{Ablation experiments on the all-in-one image restoration tasks. FA represents Fourier transform analysis, CA stands for cross-attention interaction, DCP denotes the degradation correction prompt, SP denotes strategy prompt, and DUs signifies the decision units.}
		\scalebox{0.88}{
			\begin{tabular}{c|ccccc|cc}
				\Xhline{1pt}\rule{0pt}{10pt}
				%			\toprule
				{\textbf{Method}}&{FA} &{CA} &{DCP} &{SP} & {DUs} &{PSNR↑}  & {SSIM↑}  \\
				\hline \rule{-3pt}{10pt}
				a&\ding{55} & \ding{51} & \ding{51} & \ding{51} & \ding{51}  & 31.32 & 0.880   \\
				b&\ding{51} & \ding{55} & \ding{51} & \ding{51} & \ding{51} & 31.11 & 0.876   \\
				c&\ding{55} & \ding{55} & \ding{51} & \ding{51} & \ding{51} & 30.35 & 0.865   \\
				d&\ding{51} & \ding{51} & \ding{55} & \ding{51} & \ding{51} & 31.41 & 0.872   \\
				e&\ding{51} & \ding{51} & \ding{51} & \ding{55} & \ding{51} & 31.67 & 0.884   \\
				f&\ding{51} & \ding{51} & \ding{51} & \ding{51} & \ding{55} & 31.38 & 0.892  \\
				g&\ding{51} & \ding{51} & \ding{51} & \ding{55} & \ding{55} & {30.96} & {0.875}   \\
				h&\ding{51} & \ding{51} & \ding{51} & \ding{51} & \ding{51} & {32.36} & {0.901}   \\
				%			\bottomrule
				\Xhline{1pt}
			\end{tabular}
		}
		\label{tab:ablation}
	\end{table}
	\begin{table}[t]
		\renewcommand{\arraystretch}{1.1}
		\setlength{\tabcolsep}{4.5pt}
		\small
		\centering
		\caption{Ablation experiments on task extendibility (PSNR↑), where R, H, N, B, and L represent derain, dehaze, denoise, deblur, and low-light enhancement, respectively.}
		\scalebox{0.98}{
			\begin{tabular}{c|ccccc}
				\Xhline{1pt}\rule{0pt}{10pt}
				%			\toprule
				{\textbf{Tasks}}&{Rain100L} &{SOTS} &{BSD68} & {GoPro} &{LOL}   \\
				\hline \rule{-3pt}{10pt}
				R+H+N&37.97 & 31.61  & 30.42 & 24.98  & 13.22    \\
				R+H+N+L&37.61 & 32.33  & 30.03 & 23.65 & 25.39   \\
				R+H+N+B&37.78 & 32.08  & 31.22 & 32.77 & 12.96  \\
				R+H+N+B+L&38.35 & 32.57 & 31.73 & 32.70 & 26.49  \\
				%			\bottomrule
				\Xhline{1pt}
			\end{tabular}
		}
		\label{tab:extend}
	\end{table}
	\begin{table}[t]
		\renewcommand{\arraystretch}{1.3}
		\setlength{\tabcolsep}{2.75pt}
		\footnotesize
		\centering
		\caption{Ablation experiments on the cross-domain interaction methods. $C_t$ denotes degradation correction prompt, $S_t$ represents strategy prompt, $D_f$ signifies frequency domain features, and $I$ represents spatial domain features. The best results are highlighted in~\textbf{bold}.}       
		\scalebox{1}{
			\begin{tabular*}{0.78\linewidth}{c|ccc|ccc|cc}
				\Xhline{1pt}\rule{0pt}{10pt}
				
				\multirow{2}{*}{\textbf{Method} }
				& \multicolumn{3}{c|}{$C_t$ } & \multicolumn{3}{c|}{$S_t$}  &\multicolumn{2}{c}{Average }
				\\
				
				%\cmidrule(lr){2-3}\cmidrule(lr){4-5}\cmidrule(lr){8-10}\cmidrule(lr){11-13}
				& Q  & K & V  & Q & K & V & PSNR↑  & SSIM↑  \\
				
				\hline \rule{-3pt}{10pt}
				%ProtoNet+FWT~\cite{Tseng_Lee_Huang_Yang_2020}(RN10)&-&23.77 & 26.87&-&38.87&43.78&-&67.34&75.74&-&72.72&85.82 \\
				a	&$I$ &$I$& $D_f$& $I$& $D_f$& $D_f$& 32.29& 0.897\\
				b  &$D_f$   & $D_f$ & $I$ &$I$ & $D_f$ & $D_f$&31.81 & 0.885 \\
				c	&$I$& $I$& $D_f$ &$D_f$ &$I$ &$I$ &31.58&0.891 \\
				d	&$I$ &$I$ &$D_f$& $I$ &$I$& $D_f$& 32.14& 0.895\\
				e	&$D_f$& $D_f$& $I$ & $D_f$ & $D_f$ & $I$ & 30.76& 0.873\\
				f    &$D_f$   & $D_f$ & $I$ & $D_f$ & $I$ & $I$ &\textbf{32.36}  & \textbf{0.901} \\
				\Xhline{1pt}
			\end{tabular*}
			
		}
		\label{tab:materialQKV}
	\end{table}
	
	\begin{table}[t]
		\renewcommand{\arraystretch}{1}
		\setlength{\tabcolsep}{7.3pt}
		\small
		\centering
		\caption{Ablation experiments with our dynamic decomposition mechanism. The best results are highlighted in~\textbf{bold}.}
		\scalebox{1}{
			\begin{tabular}{c|c|cc|cc}
				\Xhline{1pt}\rule{0pt}{10pt}
				%			\toprule
				{\textbf{Method}}&DDM &{PSNR↑} &{SSIM↑} &{Params}  &{FLOPs}    \\
				\hline \rule{-3pt}{10pt}
				a&\ding{55} &31.38 & 0.892 & 39.30M &41.99G    \\
				b&\ding{51} &\textbf{32.36} & \textbf{0.901}  & 37.80M & 33.67G   \\
				\Xhline{1pt}
			\end{tabular}
		}
		\label{tab:materialDDM}
	\end{table}
	
	i) When FA and CA, which together constitute CDDA, are not retained (method c), the model achieves the lowest PSNR and SSIM, underscoring the critical role of CDDA in enhancing model performance.
	
	ii) When SP and DUs are simultaneously removed (method g), the model effectively loses DDM, leading to a decline in performance, which further validates the effectiveness of DDM.

	iii) The comparisons between methods a and h, b and h, d and h, e and h, and f and h demonstrate the effectiveness of each component.
	
	iv) The complete model (method h), retaining all components, achieves the highest PSNR and SSIM, highlighting the importance of the synergistic work of all parts of the architecture.
	
	In Table~\ref{tab:extend}, we assess the scalability and comprehensive performance of the D$^3$Net model in handling multiple image degradation tasks. The results show that as the number of tasks handled increases, our method maintains stable performance and even benefits, demonstrating its robustness and scalability in handling a variety of degradation tasks.
	
	\subsubsection{Cross-Domain Interaction Methods}
	Table~\ref{tab:materialQKV} demonstrates the impact of our proposed cross-domain interaction methods on the performance of D$^3$Net. Specifically, we compare six different cross-domain interaction configurations, where $C_t$ represents degradation correction prompt, and \( S_t \) represents strategy prompt. \( D_f \) and \( I \) represent frequency domain features and spatial domain features, respectively. We observe that method f achieves the best results on both PSNR~\cite{Huynh-Thu_Ghanbari_2008} and SSIM~\cite{Wang_Bovik_Sheikh_Simoncelli_2004} metrics.
	Specifically, in the process of generating $C_t$, method f utilizes $D_f$ as the query $Q$ and key $K$, and performs a non-linear projection on $I$ to generate the value $V$. Furthermore, in the generation of $S_t$, it employs $I$ as both the query $Q$ and key $K$, and conducts a non-linear projection on $D_f$ to generate the value $V$. This method effectively integrates spatial and frequency domain features, significantly enhancing the model's performance. Notably, method a, employing a complementary configuration similar to method f, achieves the suboptimal results, further proving the importance of complementary nonlinear projection of frequency and spatial domain features at the $V$ and $Q$ positions in degradation correction prompt and strategy prompt. The ablation study's results indicate that effectively integrating spatial and frequency domain features via method f leads to significant enhancements in image quality, highlighting the importance of the specific cross-domain interaction configuration in improving model performance.
		\begin{table}[t]
		\renewcommand{\arraystretch}{1.3}
		\setlength{\tabcolsep}{18pt}
		\small
		\centering
		\caption{Impact of DUs' number on PSNR↑ and SSIM↑.}
		\scalebox{1}{
			\begin{tabular}{c|cc}
				\Xhline{1pt}\rule{0pt}{8pt}
				%	\toprule 
				{\textbf{Number of DUs}} &{PSNR↑} &{SSIM↑}     \\
				%			\hline \rule{-3pt}{9pt}
				\hline \rule{-3pt}{7pt}
				3 &31.73 & 0.894    \\
				6 &32.14 & 0.899    \\
				9 &32.27 & 0.897   \\
				12 &\textbf{32.36} & \textbf{0.901}    \\
				14 &32.31 & 0.896    \\
				
				%			\bottomrule	
				\Xhline{1pt}
			\end{tabular}
		}
		\label{tab:DUs}
	\end{table}
	\begin{figure}[t] 
		\centering 
		\includegraphics[width=0.49\textwidth]{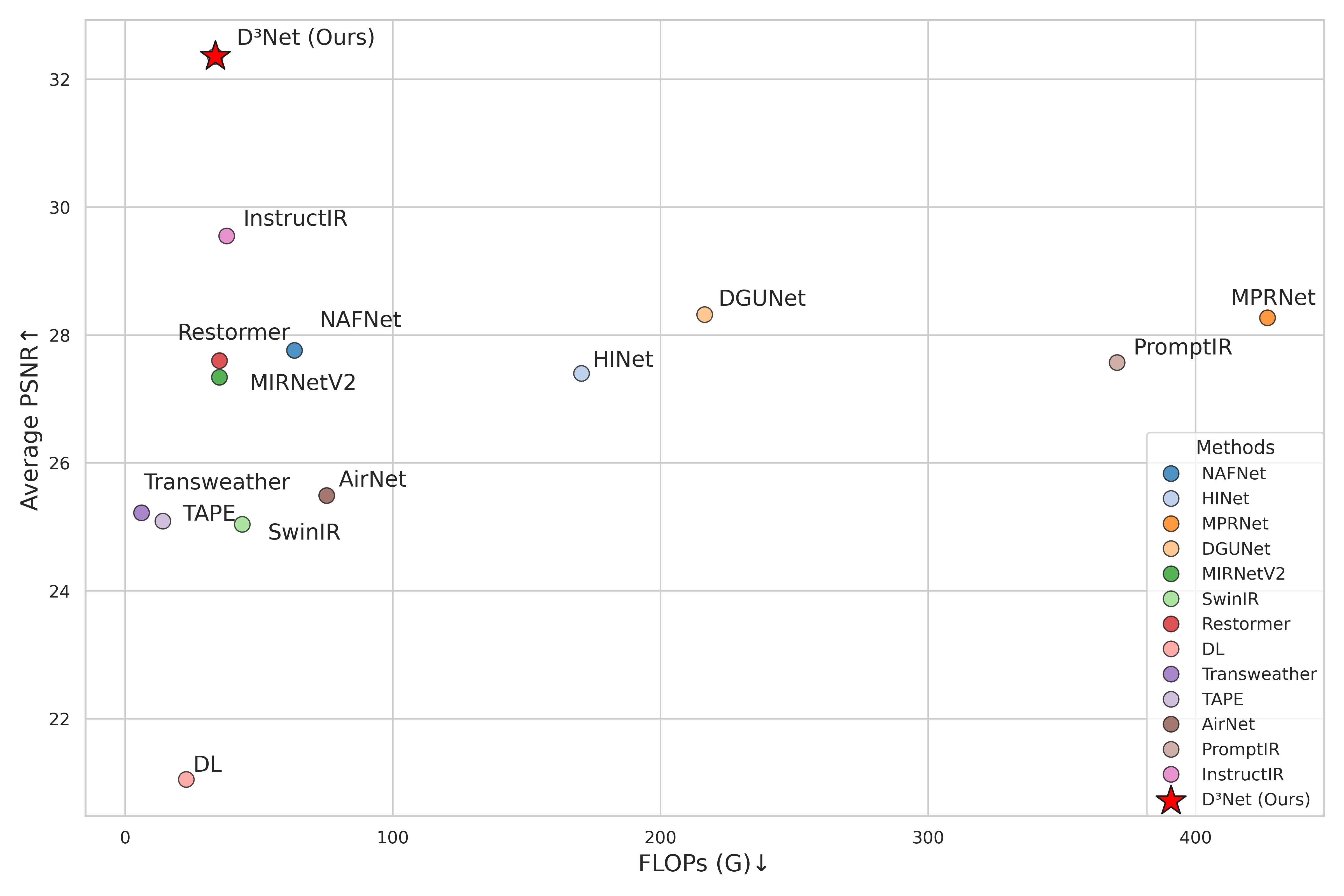} 
		\caption{Performance comparison between D$^3$Net and the SOTA methods on all-in-one image restoration tasks.} 
		\label{fig:Flops} 
	\end{figure}
	\begin{figure}[t]
		\centering
		%	\fbox{\rule{0pt}{0.5in} \rule{0.9\linewidth}{0pt}}43
		\includegraphics[width=0.5\textwidth]{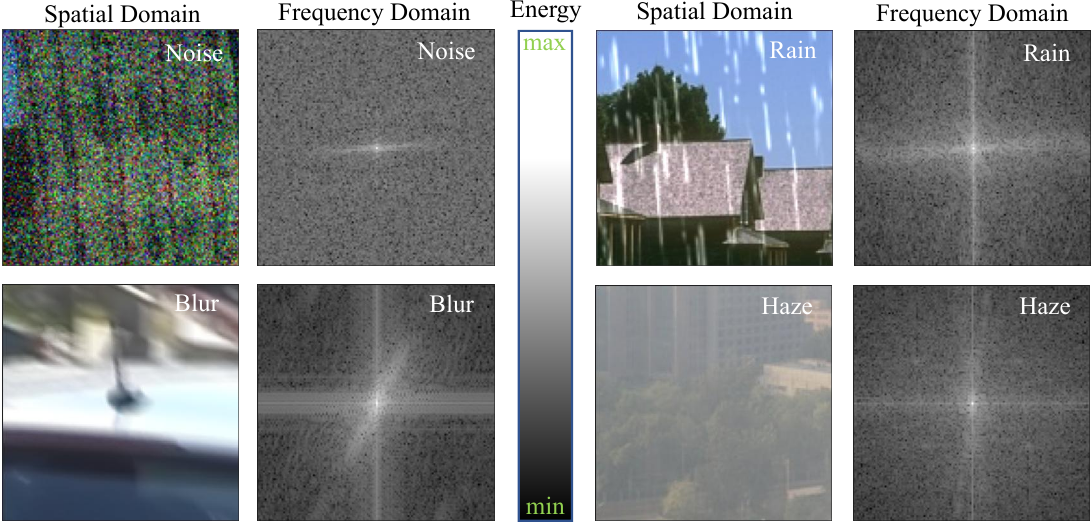}
		\caption{Degradation impact on frequency domain features.}
		\label{fig:Vis}
	\end{figure}
	\begin{figure}[t] 
		\centering 
		\includegraphics[width=0.5\textwidth]{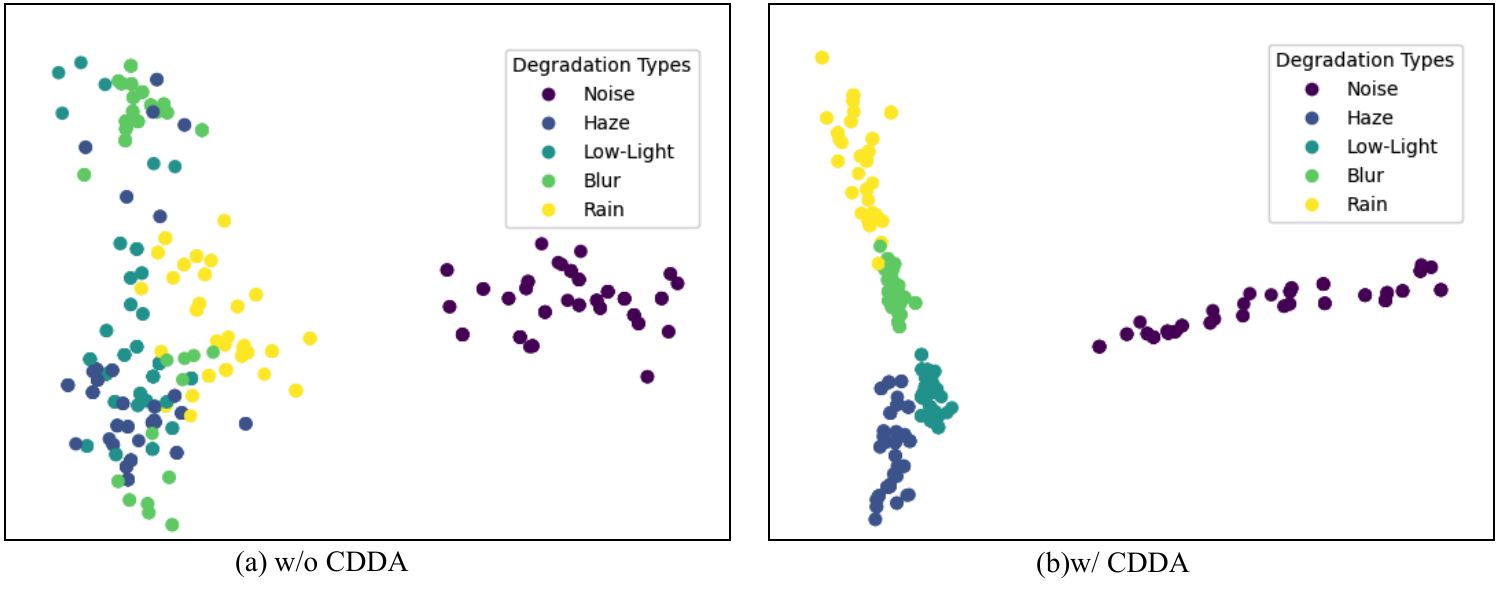} 
		\caption{Impact of CDDA on the visualization of degradation representations.} 
		\label{fig:tSNE} 
	\end{figure}
	\subsubsection{Dynamic Decomposition Mechanism}
	To validate the effectiveness of our proposed dynamic decomposition mechanism in all-in-one image restoration tasks, we perform ablation experiments. The results of these experiments, detailed in Table~\ref{tab:materialDDM}, present the influence of DDM on the efficacy of the overall restoration performance. It demonstrates that our DDM improves D$^3$Net's PSNR and SSIM while reducing the number of parameters and computational overhead.
	%It demonstrates the performance of our D$^3$Net model in terms of image quality metrics PSNR and SSIM, and the model's Params and FLOPs, with and without DDM.
	Specifically, method b, compared to method a, shows an improvement of 0.98dB in PSNR and 0.009 in SSIM, while reducing the computational overhead by \( 19.81\% \). This result adequately demonstrates that D$^3$Net, through the utilization of DDM, can achieve better image restoration results while reducing redundant computational expenses.
	
	\subsubsection{Decision Units}
	To validate the effectiveness of the proposed decision units (DUs) in image restoration, we conduct ablation studies on the number of DUs within the D$^3$Net architecture. Table~\ref{tab:DUs} shows that increasing the number of DUs generally improves both average PSNR and SSIM across the five evaluated image restoration tasks, confirming the benefit of multiple decision units. Performance peaks at 12 DUs, with PSNR reaching 32.36 and SSIM reaching 0.901. A slight performance decrease is observed at 14 DUs, suggesting diminishing returns and a potential increase in model complexity without a corresponding performance increase. Consequently, 12 DUs are selected for subsequent experiments, representing an optimal balance between performance and model complexity.
	\subsection{Experimental Investigation of D³Net's Frequency-Aware Degradation Modeling and Restoration} 
	%		In this section, we perform empirical experiments to visually investigate and elaborate on our research findings.
	In this section, we conduct a series of empirical experiments to investigate and elucidate D$^3$Net's frequency-domain-aware approach to modeling and restoring various image degradations. 
	%In this section, we conduct some visualization experiments and extended discussions.
	
	As shown in Fig.~\ref{fig:Flops}, we present a performance comparison of D$^3$Net with other SOTA image restoration methods~\cite{Li2022AllInOneIR,PromptIR_NIPS2024,Jose_Valanarasu_Yasarla_Patel_2022,Liu2022TAPETP,Chen_Chu_Zhang_Sun_2022,Zamir_Arora_Khan_Hayat_Khan_Yang_Shao_2021,Zamir2022LearningEF,Zamir_Arora_Khan_Hayat_Khan_Yang_2022,Fan_Chen_Yuan_Hua_Yu_Chen_2021,Liang2021SwinIRIR,Chen_Lu_Zhang_Chu_Chen_2021,Mou_Wang_Zhang_2022}, where PSNR represents the average PSNR value across multiple image restoration tasks. Compared to other methods, D$^3$Net achieves excellent performance under reasonable computational overhead.

	In addition, from the perspective of frequency domain, we investigate the impact of different types of degradation on the frequency domain of images, as shown in Fig.~\ref{fig:Vis}. Noise typically appears as high-frequency components in the frequency domain, while image blur results in a reduction of high-frequency components. Raindrops lead to horizontal or diagonal stripes, and haze generally causes a decrease in low-frequency components. Inspired by these observations, we propose a cross-domain degradation analyzer. It deeply integrates frequency domain degradation features with spatial domain image features, generating degradation correction prompt and strategy prompt. These prompts offer effective guidance for the dynamic decomposition mechanism, enabling D$^3$Net to employ more specific and effective restoration strategies for each type of degradation.
	
	Furthermore, we explore the impact of CDDA on the performance of D$^3$Net for different degradation types. As shown in Fig.~\ref{fig:tSNE}, we employ t-SNE~\cite{Maaten_Hinton_2008} to visualize the decomposition feature of D$^3$Net by taking it as input for five different types of degradation representations: noise, haze, low-light, blur, and rain. Fig.~\ref{fig:tSNE}(a) shows the distribution of degradation  representations without CDDA, while Fig.~\ref{fig:tSNE}(b) shows the situation with CDDA.
	From Fig.~\ref{fig:tSNE}(a), it is observed that without CDDA, the boundaries between different degradation types are not distinct, with overlapping representation points, indicating the challenges in distinguishing different degradation types under these conditions. However, in Fig.~\ref{fig:tSNE}(b), each degradation type exhibits a more distinct and segregated distribution, indicating that the CDDA significantly enhances the model's ability to recognize different degradation types. 
	\section{Conclusion}
	In this paper, we propose a novel all-in-one image restoration network, termed D$^3$Net, as a comprehensive architecture capable of handling multiple image restoration tasks. Distinct from existing all-in-one image restoration methods, the uniqueness of D$^3$Net lies in its ability to guide the network learning process by deeply interacting frequency domain degradation characteristics with spatial domain image features to generate degradation correction prompt and strategy prompt, endowing the model with degradation adaptability. Moreover, we introduce a prompt-based dynamic decomposition mechanism for adaptable degradation decomposition. Owing to its adaptive prompt-guided mode, this mechanism can achieve superior degradation decomposition effects while reducing computational overhead. Extensive experiments demonstrate the effectiveness and scalability of the proposed D$^3$Net.

	Considering the reliance of the dynamic decomposition mechanism on the quality of prompt, we will explore a more robust and accurate mechanism for generating prompt in the future.
	\bibliographystyle{IEEEtran}
	\bibliography{main}
	
	%		\vspace{15pt}
	%	%	

\end{document}